\documentclass[conference]{IEEEtran}

\usepackage{cite}

\usepackage[pdftex]{graphicx}
\usepackage{caption}
\usepackage{color}
\usepackage{subcaption}
\usepackage{amsmath}
\usepackage[nolist,nohyperlinks]{acronym}
\usepackage{soul}
\ifCLASSINFOpdf
\else
\fi
\usepackage{amsmath}

\usepackage[cmintegrals]{newtxmath}
\begin{document}
%
\title{ML Approach for Power Consumption Prediction in Virtualized Base Stations}
\author{
	Merim Dzaferagic\IEEEauthorrefmark{1}, Jose A. Ayala-Romero\IEEEauthorrefmark{2} and Marco Ruffini\IEEEauthorrefmark{1}\\
	\IEEEauthorblockA{\IEEEauthorrefmark{1}CONNECT Research Centre, School of Computer Science and Statistics, Trinity College Dublin\\}
	\IEEEauthorblockA{\IEEEauthorrefmark{2}NEC Laboratories Europe GmbH\\}
	Emails: merim.dzaferagic@tcd.ie, jose.ayala@neclab.eu and marco.ruffini@tcd.ie
	\thanks{Merim Dzaferagic and Marco Ruffini are with the CONNECT Research Centre, School of Computer Science and Statistics, Trinity College Dublin.}
	\thanks{Jose A. Ayala-Romero is with NEC Laboratories Europe GmbH}  
	\thanks{We would like to say thank you to Diarmuid Collins for the help in the use and discussions related to srsRAN.}
	\thanks{This material is based upon works supported by the Science Foundation Ireland under Grants No. 17/CDA/4760 and 13/RC/2077\_P2.}\vspace{-1em}
}
%

%



\maketitle

\begin{abstract}
The flexibility introduced with the Open Radio Access Network (O-RAN) architecture allows us to think beyond static configurations in all parts of the network. This paper addresses the issue related to predicting the power consumption of different radio schedulers, and the potential offered by O-RAN to collect data, train models, and deploy policies to control the power consumption. We propose a black-box (Neural Network) model to learn the power consumption function. We compare our approach with a known hand-crafted solution based on domain knowledge. Our solution reaches similar performance without any previous knowledge of the application and provides more flexibility in scenarios where the system behavior is not well understood or the domain knowledge is not available.


\end{abstract}



%
\IEEEpeerreviewmaketitle

\begin{acronym}
\acro{ai}[AI]{Artificial Intelligence}
\acro{bbu}[BBU]{Baseband Unit}
\acro{pcp}[PCP]{Power Consumption Prediction}
\acro{smo}[SMO]{Service Management and Orchestration}
\acro{non-rt-ric}[non-RT RIC]{non-Real Time RAN Intelligent Controller}
\acro{cots}[COTS]{Commercial off-the-shelf}
\acro{rru}[RRU]{Radio Remote Unit}
\acro{rrh}[RRH]{Remote Radio Head}
\acro{bs}[BS]{Base Station}
\acro{vbs}[vBS]{Virtual Base Station}
\acro{mcs}[MCS]{Modulation and Coding Scheme}
\acro{cqi}[CQI]{Channel Quality Indicator}
\acro{rmse}[RMSE]{Root Mean Squared Error}
\acro{nn}[NN]{Neural Network}
\acro{snr}[SNR]{Signal to Noise Ratio}
\acro{lsm}[LSM]{Least Squares Method}
\acro{oran}[O-RAN]{Open Radio Access Network}
\acro{ran}[RAN]{Radio Access Network}
\acro{cran}[C-RAN]{Cloud Radio Access Network}
\acro{vran}[vRAN]{Virtualized Radio Access Network}
\acro{iot}[IoT]{Internet of Things}
\acro{bs}[BS]{Base Station}
\end{acronym}
\section{Introduction}
The increasing number of diverse devices (e.g. mobile users, \ac{iot} devices, vehicles) in mobile networks inevitably leads to an increasing need for up-scaling the network capacity. We are already witnessing the efforts that mobile operators are putting into densifying the network by deploying more \acp{bs} and sharing the existing resources. Additionally, the existing network architectures are not built with sufficient flexibility and intelligence to efficiently handle the increasing resource demands \cite{saad2019vision}. An architectural transformation is required to support the heterogeneous device and service demands, on-demand service deployment, security updates and coordination of multi-connectivity technologies \cite{niknam2020intelligent}. \ac{oran}, with its main goal to enhance the \ac{ran} through virtualization of the network elements and open interfaces that incorporate intelligence, emerged as a solution for all the abovementioned issues \cite{oran2018}.

As highlighted by the authors of \cite{niknam2020intelligent}, unlike \acp{cran} and \acp{vran}, \acp{oran} uses well defined open interfaces between the elements implemented in general-purpose hardware. \ac{oran} also allows \ac{rru} and \ac{bbu} hardware and software from different vendors. All this flexibility makes it hard to optimize the operation of the softwarized entities on the \ac{cots} hardware. This also leads to increased energy consumption, which already is the biggest expense for the operators and is expected to further grow \cite{gsma_energy}. Considering that the telecoms industry consumes between $2-3\%$ of global energy, energy consumption is not just an issue from the environmental point of view, but also a big liability for telecom operators. The approaches to solving this problem range from using sustainable energy sources \cite{chamola2016solar} to the development of energy efficient network protocols and algorithms \cite{zhang2017energy}.

One promising approach in new generation networks is the \textit{network virtualization}, whereby the BS functions are implemented in software and deployed in general-purpose CPU or shared computational pools.
In virtualized networks, the \ac{bbu} becomes a predominant contributor in the consumed power and its computational load (and hence its consumed power) depends on several parameters in different ways.
Previous works show that the relation between the operational configuration of the \acp{vbs} is not straightforward presenting non-linear relations as shown in \cite{rost2015complexity, ayala2021experimental}. 

In this paper, we study how to predict the consumed power associated with different radio schedulers of \acp{vbs}. More precisely, we propose a black-box \ac{nn} approach and compare it with the solution proposed in \cite{ayala2021experimental}, which relies on domain knowledge.

Extensive research has been done on estimating the \textit{power consumption of legacy \acp{bs}} \cite{2013SohEnergy,surveyBuzzi2016,joung2014survey,mahmood2015modeling}. The work in \cite{2013SohEnergy} studies the energy efficiency in homogeneous macrocell and heterogeneous K-tier wireless networks under different sleeping policies. They also formulate a power consumption minimization problem to determine the optimal operating regimes for macrocell \acp{bs}. The authors of \cite{surveyBuzzi2016} highlight the importance of designing energy-efficient hardware solutions (e.g. green design of the radio frequency chain, the use of simplified transmitter/receiver structures). In other works like \cite{joung2014survey, mahmood2015modeling}, the focus is on the energy-efficient design of power amplifiers.

In contrast to legacy \acp{bs}, the authors of \cite{CaoEnergy2018} highlight that the total power consumption in \acp{vbs} consists of the baseband signal processing power and the radio power. They further present evidence proving that a \textit{computation-resource-aware approach over \acp{vbs}} (i.e. an approach that jointly optimizes the data rate and the number of CPU cores) can save more than 60\% of energy compared to legacy \acp{bs}. 
Based on these ideas, other follow-up works design energy-efficient configuration policies taking into account user performance constraints \cite{ayala2021orchestrating} and edge AI service QoS requirements \cite{ayala2021edgebol}.

In contrast to the domain knowledge-based solution in \cite{ayala2021experimental}, we present a black-box model that does not consider any previous knowledge of the application. Moreover, our solution can be easily extended and trained to consider new dimensions (e.g., the bandwidth), while the hand-crafted solution in \cite{ayala2021experimental} becomes very difficult to modify and adapt to new scenarios.
Finally, we also show how to automate the process of data collection and model training in the \ac{oran} architecture.

\section{Approximating the power function }\label{sec:data}

Our goal is to approximate the power behavior of a \ac{vbs} scheduler with a black-box model (\ac{nn}) as a function of the parameters with a predominant importance in the consumed power, i.e., airtime, \ac{mcs}, \ac{snr} \cite{ayala2021experimental, ayala2021orchestrating}. We are also interested in testing whether domain knowledge needs to be embedded in the design of a model to accurately approximate the power function of a generic scheduler. Therefore, we compare the performance of our \ac{nn} to a known regression model used on the same dataset.  

The two steps involved in the process of building a black-box model and comparing it to a regression model are: (1) data collection; (2) choosing and training a model.

\subsection{Data collection}\label{sub:data_collection}

\begin{figure}[h]
     \centering
     \begin{subfigure}[b]{0.24\textwidth}
         \includegraphics[trim={0 0 0 0},clip,scale=0.30]{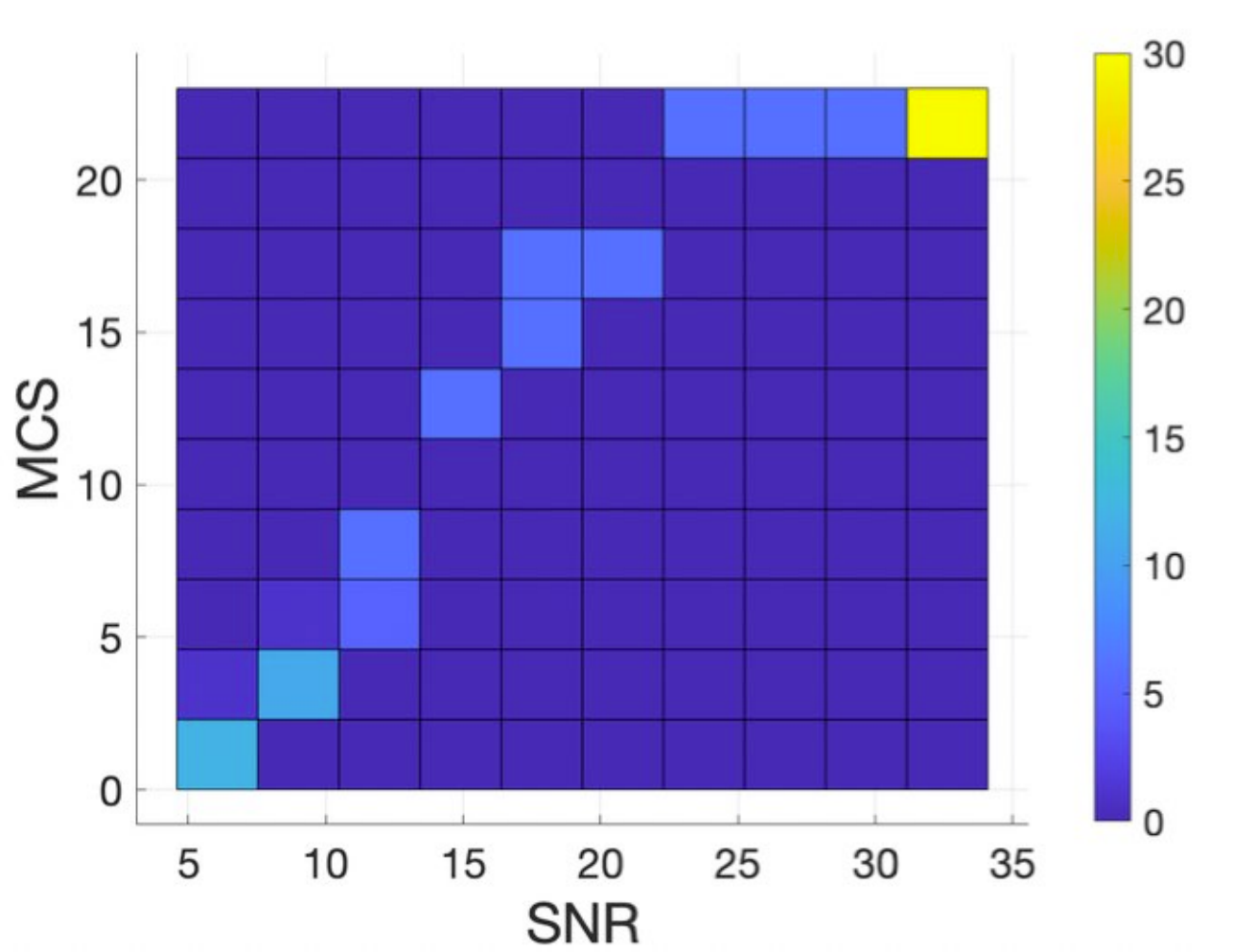}
         \caption{}\label{fig:scheduler_only_dataset_server1}
     \end{subfigure}
     \begin{subfigure}[b]{0.24\textwidth}
         \includegraphics[trim={0 0 0 0},clip,scale=0.30]{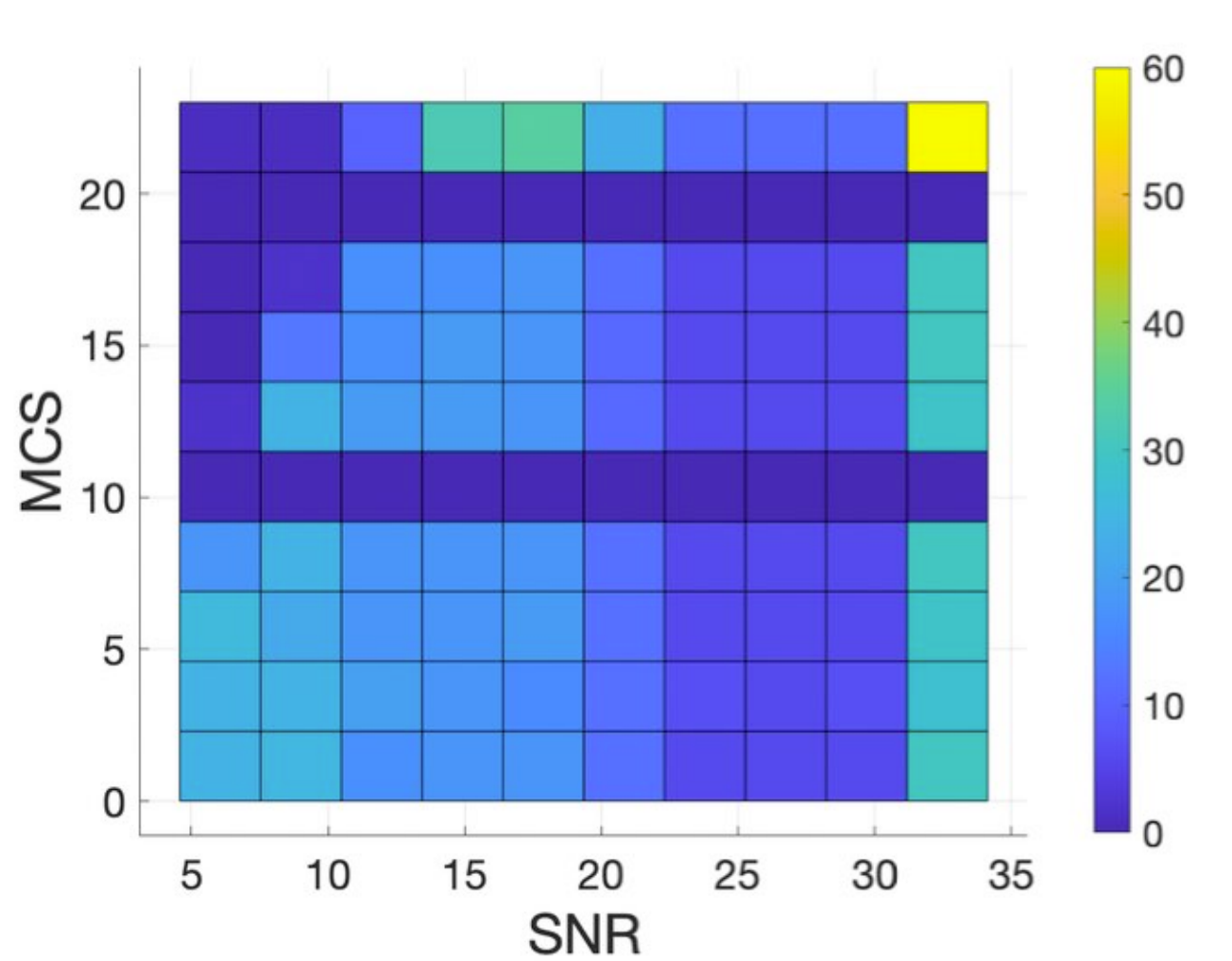}
         \caption{}\label{fig:extended_dataset_server1}
     \end{subfigure}
        \caption{\textcolor{black}{Default and custom scheduler dataset samples for Server 1. The colors indicate the number of samples for each combination of SNR and MCS - darker colors indicate less samples.}}
        \label{fig:dataset_visualization}
        \vspace{-1em}
\end{figure}

The first step in the process of building a model to predict the power consumption of a generic scheduler is to collect a dataset that will be used for training and testing the performance of those models. In this paper, we will use the dataset collected by the authors of \cite{ayala2021experimental}. The dataset contains \ac{vbs} power consumption measurements. The authors measure the power via software and hardware. The software measurements were collected by using the Intel Running Average Power Limit functionality integrated into the Linux kernel to measure the CPU power. The hardware measurements were obtained with the GPM-8213 meter connected to the GPM-001 measuring adapter, which provides power to the \ac{bs} through a power supply cable. The \ac{vbs} consists of a \ac{rrh} for which they use the Ettus Research USRP B210, and a \ac{bbu}. The authors used four different computing platforms for the \acp{bbu} (see Table~\ref{tab:computing_platforms}), and a customized version of \textit{srsRAN} \cite{gomez2016srslte}, which allows them to change the \ac{mcs} and airtime through a TCP socket on the fly. This allows them to collect two datasets shown in Fig.~\ref{fig:dataset_visualization} referred to as: (1) \textit{default scheduler dataset}; and (2) \textit{custom scheduler dataset}.
\vspace{-1em}

Fig.~\ref{fig:scheduler_only_dataset_server1} shows the distribution of the samples in the default scheduler dataset, which contains power measurements for the scenario in which the default \textit{srsRAN} radio scheduler selects the \ac{mcs} for each given measured channel quality. We can observe that the higher the channel quality, the higher the selected \ac{mcs}. These power consumption samples can be obtained during the normal operation of the default scheduler in an operational network. However, Fig.~\ref{fig:scheduler_only_dataset_server1} also shows that a large part of the possible combinations of \ac{snr} and \ac{mcs} are not selected and therefore are missing in this dataset. The custom scheduler dataset, on the other hand, was collected by modifying the default scheduler, i.e.,  different \ac{mcs} are selected for the same measured channel quality (see Fig.~\ref{fig:extended_dataset_server1}). 
The acquisition of this measurements is very risky for an operational \acs{vbs} because there are some combinations of \ac{snr} and \ac{mcs} that are not feasible resulting in decoding errors (see top left corner in Fig.~\ref{fig:extended_dataset_server1}). These configurations have a low channel quality and a high \ac{mcs} so that the signal is too noisy to be decoded.
 However, the custom scheduler dataset allows us to study the relationship between the input features in the whole search space, potentially allowing us to design a new power consumption-aware scheduler. The \ac{oran} architecture would allow us to deploy a modified scheduler dynamically, depending on the network needs. 

\begin{figure*}[h]
  \centering
  \includegraphics[scale=0.35]{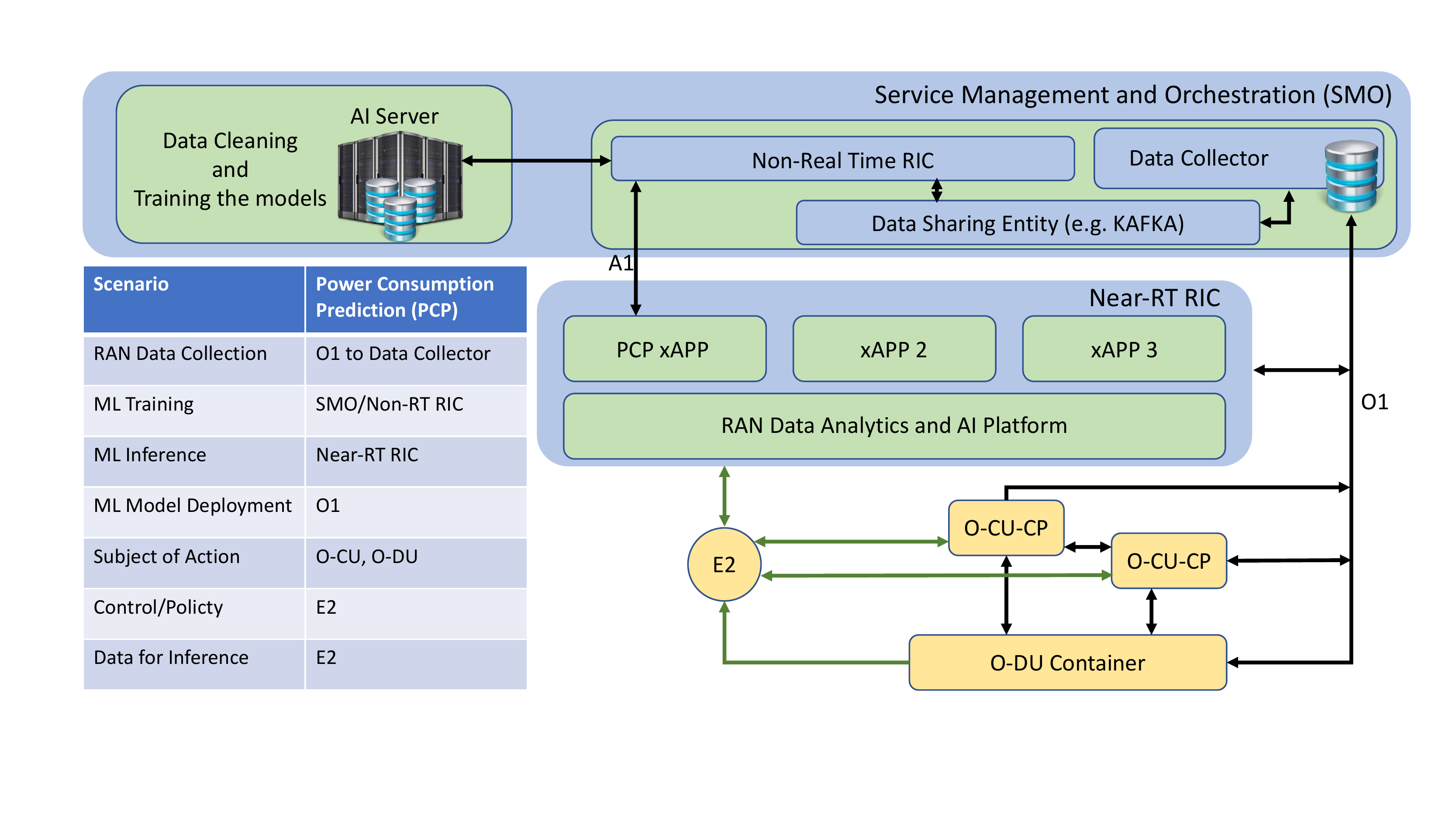}
    \vspace{-9mm}
  \caption{\ac{oran} deployment architecture.}
  \label{fig:oran_deployment_architecture}
    \vspace{-5mm}
\end{figure*}


\begin{table}[h]
\caption{Computing platforms used for the \acp{bbu}.}\label{tab:computing_platforms}
\begin{tabular}{l|ccll}
\textit{Alias} & Commercial name          & CPU                &  &  \\ \hline
NUC1           & BOXNUC8I7BEH             & i7-8559U @ 2.70GHz &  &  \\
NUC2           & NUC7i7DNHE               & i7-8650U @ 1.90GHz &  &  \\
Server1        & Dell XPS 8900 Series     & i7-6700 @ 3.40GHz  &  &  \\
Server2        & Dell Alienware Aurora R5 & i7-9700 @ 3.00GHz  &  & 
\end{tabular}
\vspace{-1em}
\end{table}
 
The evaluation will investigate how a black-box model (\ac{nn}) compares to a regression model with domain knowledge built into it. We compare the two proposed models based on the \ac{rmse} of the predicted power consumption for each of the computing platforms shown in Table~\ref{tab:computing_platforms}. 

The next steps involve choosing, training, and comparing the performance of the chosen models. We will start with the regression model proposed by the authors of \cite{ayala2021experimental}. Then, we will propose a black-box (\ac{nn}) model and train it to approximate the power behavior of the \acp{vbs}.

\subsection{Regression models}\label{subsec:regression_models_intro}
The authors of \cite{ayala2021experimental} propose two regression models: (1) a model designed for default scheduler data; and (2) a model designed for the custom scheduler dataset. 
\subsubsection{Default scheduler dataset} Let us introduce the first model by computing the CPU power consumption $P$ with: 
\begin{equation}
  P(a,c) =\begin{cases} P_0(a)-r(a)\cdot(\gamma_0-c) & c < \gamma_0 \\
                     P_0(a) &  \text{otherwise},
       \end{cases}  
\end{equation}
where $c\in\mathbb{R}^+$ denotes the \ac{snr} in dBs, and $a\in[0,1]$ the airtime, where $a = 1$ indicates that all subframes are used, and $a = 0$ indicates zero throughput. $P_0(a)$ and $r(a)$ are the maximum power for a fixed airtime value and slope of the power consumption curve, respectively, which are defined as: 
\begin{equation}
    P_0(a) = \gamma_1 + \gamma_2\cdot a,\;\;\;\;\;\;r(a) = \gamma_3 + \gamma_4\cdot a.
\end{equation}
The values of $\gamma = (\gamma_0,..., \gamma_4)$ vary depending on the bandwidth and the computing platform, and are obtained using the \ac{lsm} in the dataset. 

\subsubsection{Custom Scheduler dataset}The previous model shows the relationship between the power consumption and the airtime and \ac{mcs} selected by the scheduler. For the second model, we have to explore combinations that are never chosen by the default scheduler. However, the authors of \cite{ayala2021experimental}, highlight that for each \ac{mcs} $m\in\mathbb{Z}$ there is an SNR $c_{\text{th}}(m)$ below which the computational load increases. Since the slope of the power consumption depends on the airtime and \ac{mcs}, they also model the increase in power consumption with $r(a,m)$. Considering all this, the second model, which is designed for the custom scheduler dataset, is computed by: 
\begin{equation}\label{eq:regression_model_2}
  P(a,c,m)=\begin{cases} P_0(a,m)+r(a,m)\cdot c & \text{if } c_{\text{min}}(m)<c<c_{\text{th}}(m) \\
                     P_0(a,m) &  \text{if } c>c_{\text{th}}(m)
       \end{cases}  
\end{equation}
$P_0(a,m)$ is computed as follows: 
\begin{equation}
    P_0(a,m)=\beta_0+\beta_1\cdot a+\beta_2\cdot m+\beta_3\cdot a^2+\beta_4\cdot m^2+\beta_5\cdot a\cdot m
\end{equation}
\begin{equation}
    r(a,m)=\beta_6+\beta_7\cdot a+\beta_8 m
\end{equation}
\begin{equation}
    c_{\text{min}}(m)=\beta_9+\beta_{10}\cdot m
\end{equation}
\begin{equation}
    c_{\text{th}}(m)=\beta_{11}+\beta_{12}\cdot m
\end{equation}
The authors also highlight that $P(a,c,m)$ is not defined for $c<c_{min}(m)$, because the combination of $m$ and $c$ is not feasible in that case. Additionally, the values of $\beta=(\beta_0, ...,\beta_{12})$ are fitted using \ac{lsm}, and similar to the $\gamma$-values for the previous model, depend on the computing platform and radio bandwidth. 

\subsection{Black-box model (\ac{nn})}
Unlike the regression model, which was designed based on domain knowledge, the black-box (\ac{nn}) model is a generic model with no information about the problem at hand. The design process of such models relies on standard procedures in data science (e.g. data collection-data preparation-model selection-model training-testing and visualization). An obvious advantage of a black-box approach is that no specialist knowledge about the problem at hand is needed to build a model capable of predicting the consumed power based on the chosen input features. However, usually such models require a lot of data to be trained. 

Fortunately, considering the previously presented regression models, the power function does not seem to be very complex. Therefore, a small \ac{nn} can be considered to accurately model the problem. We propose a \ac{nn} with:
\begin{itemize}
    \item an input layer with three features (airtime, \ac{snr} and \ac{mcs});
    \item two hidden layers (first hidden layer with $24$ and second hidden layer with $4$ nodes, all using the \textit{ReLU} activation function);
    \item an output layer with one node that uses the \textit{ReLU} activation function.
\end{itemize}

During training, we had to consider the small size of the dataset (see Table~\ref{tab:training_testing_dataset}). Hence, the \ac{nn} was trained with the Adam optimiser with a batch size of $32$ samples and a constant learning rate $0.001$ for $220$ epochs. Considering the sparsity of the dataset (see Fig.~\ref{fig:dataset_visualization}), we introduced activity regularization \textit{l1}, to reduce the generalization error (i.e. avoid overfitting).

\subsection{Deployment Architecture}\label{sub:deployment_architecture}

Fig.~\ref{fig:oran_deployment_architecture} depicts the \ac{oran} deployment architecture for the \ac{pcp} scenario. Similar to the work presented in \cite{niknam2020intelligent}, in which the authors map the congestion prediction and mitigation scenario on the \ac{oran} architecture, we perform a mapping for our models presented in the previous sections. The goal is to show that the process of data collection and cleaning, model training and deployment, and in the end the inference itself can be automated in the \ac{oran} architecture. 

The data collection involves the collection of \ac{ran} counters from the control and distributed units through the \textit{O1} interface. The data is stored in the data collector that is located in the \ac{smo}. Please notice that for the purpose of power measurements this can become an issue in a live network. The process is an intrusive method that requires user isolation and might result in network performance degradation. However, as we will show in the results section this process is not very important since data collected offline (e.g. data collected in a laboratory environment) is enough to train the models for different computing platforms. 

The data collected from the network through the \textit{O1} interface or data collected offline is shared with the \ac{non-rt-ric} through a data sharing entity. All of the entities are located in the \ac{smo}. The \ac{ai} servers are used for data cleaning and training. After the training is finalized, the trained model is forwarded through the \ac{non-rt-ric} over the \textit{A1} interface to the \ac{pcp} xAPP.  The xAPP predicts the CPU power consumption of \acp{vbs} in one area and adjusts the scheduler behavior according to preferred policies (e.g. low power consumption, high throughput). Finally, the selected scheduler behavior is applied to the DU through the \textit{E2} interface. 

\section{Evaluation}\label{sec:results}
In this section, we evaluate the performance of the two models introduced in the previous section. We will divide the evaluation into three subsections: 
\begin{itemize}
    \item Default scheduler dataset comparing the performance of the regression model that was built for the default scheduler dataset and our black-box (\ac{nn}) model;
    \item Mixed dataset that compares how well the two models that are trained for a specific type of scheduler generalize and enable the prediction of the power consumption for any generic scheduler;
    \item Custom scheduler dataset comparing the performance of the two models in case of data from the generic scheduler being available for training. 
\end{itemize}

It is important to notice that in our evaluation we use two regression models presented in Section~\ref{subsec:regression_models_intro} (the one designed for default scheduler data is used for the evaluation in Section~\ref{subsec:scheduler_only_results}, and the more complex one is used in Section~\ref{subsec:mixed_scheduler-only_and_extended_results} and \ref{sub:extended_dataset_results}).

\subsection{Default Scheduler Dataset}\label{subsec:scheduler_only_results}

As mentioned in Section \ref{sub:data_collection}, the default scheduler dataset contains power measurements for the operation of the default \textit{srsRAN} radio scheduler. Considering that the \textit{srsRAN} scheduler chooses the \ac{mcs} of each user based on the \ac{cqi} and the mapping between the \ac{cqi} and the maximum code rate, this dataset includes only the power measurements for the combinations of measured channel quality and \ac{mcs} that are allowed based on this mapping. For example, the scheduler will never choose a high \ac{mcs} in case the channel quality is low. 

The evaluation in this section investigates the \ac{rmse} of predicting the power consumption of the default scheduler based on the collected dataset. We divide the dataset for each of the used computing platforms into a training and testing set according to the $80/20$ split (i.e., $80\%$ training data and $20\%$ testing data). The number of samples in the training and testing set for each computing platform is shown in Table~\ref{tab:training_testing_dataset}. 

\begin{figure}[t]
     \centering
     \begin{subfigure}[b]{0.24\textwidth}
         \includegraphics[trim={0 0 0 0},clip,scale=0.3]{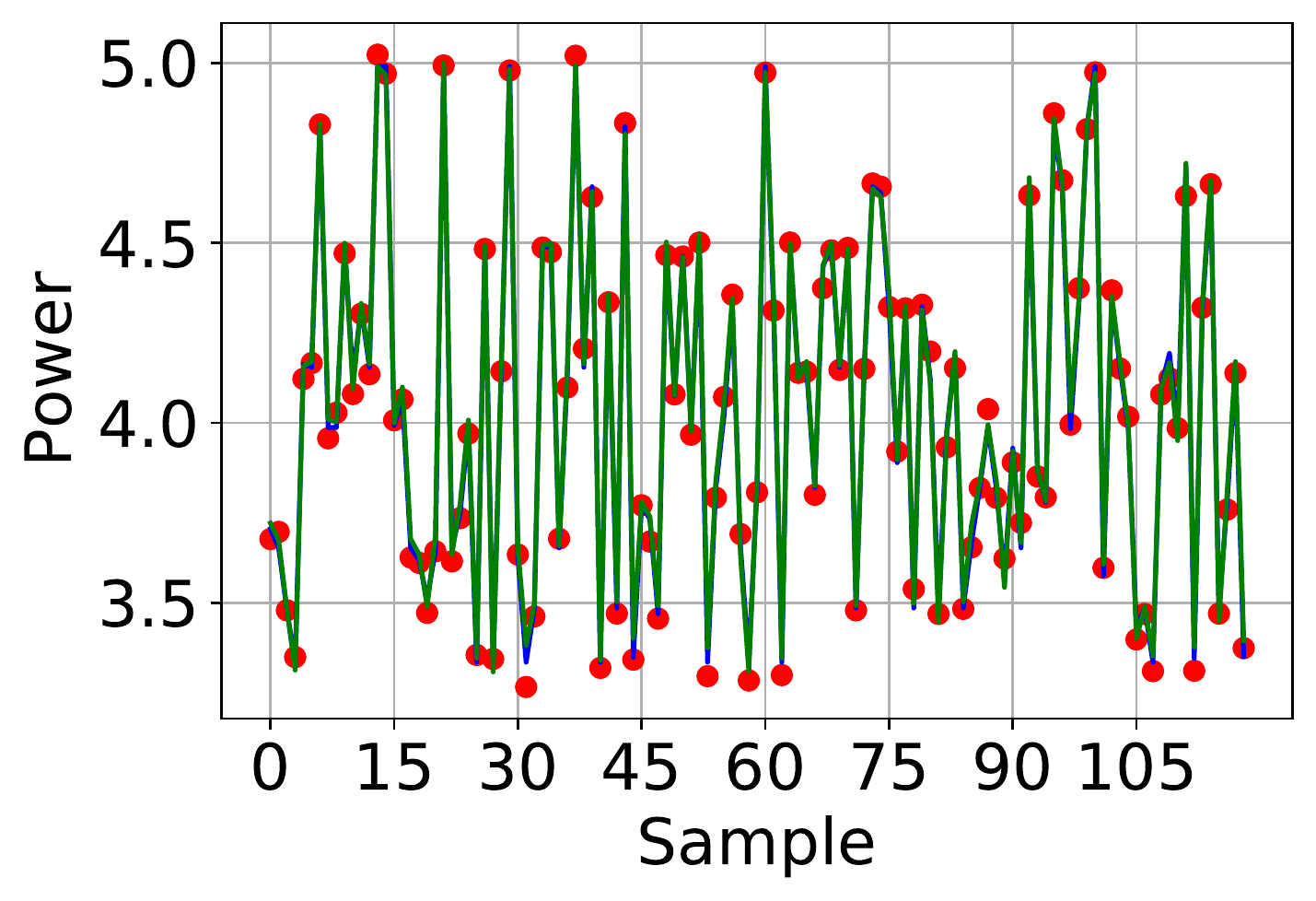}
         \caption{}\label{fig:scheduler_only_prediction_NUC1}
     \end{subfigure}
     \begin{subfigure}[b]{0.24\textwidth}
         \includegraphics[trim={0 0 0 0},clip,scale=0.3]{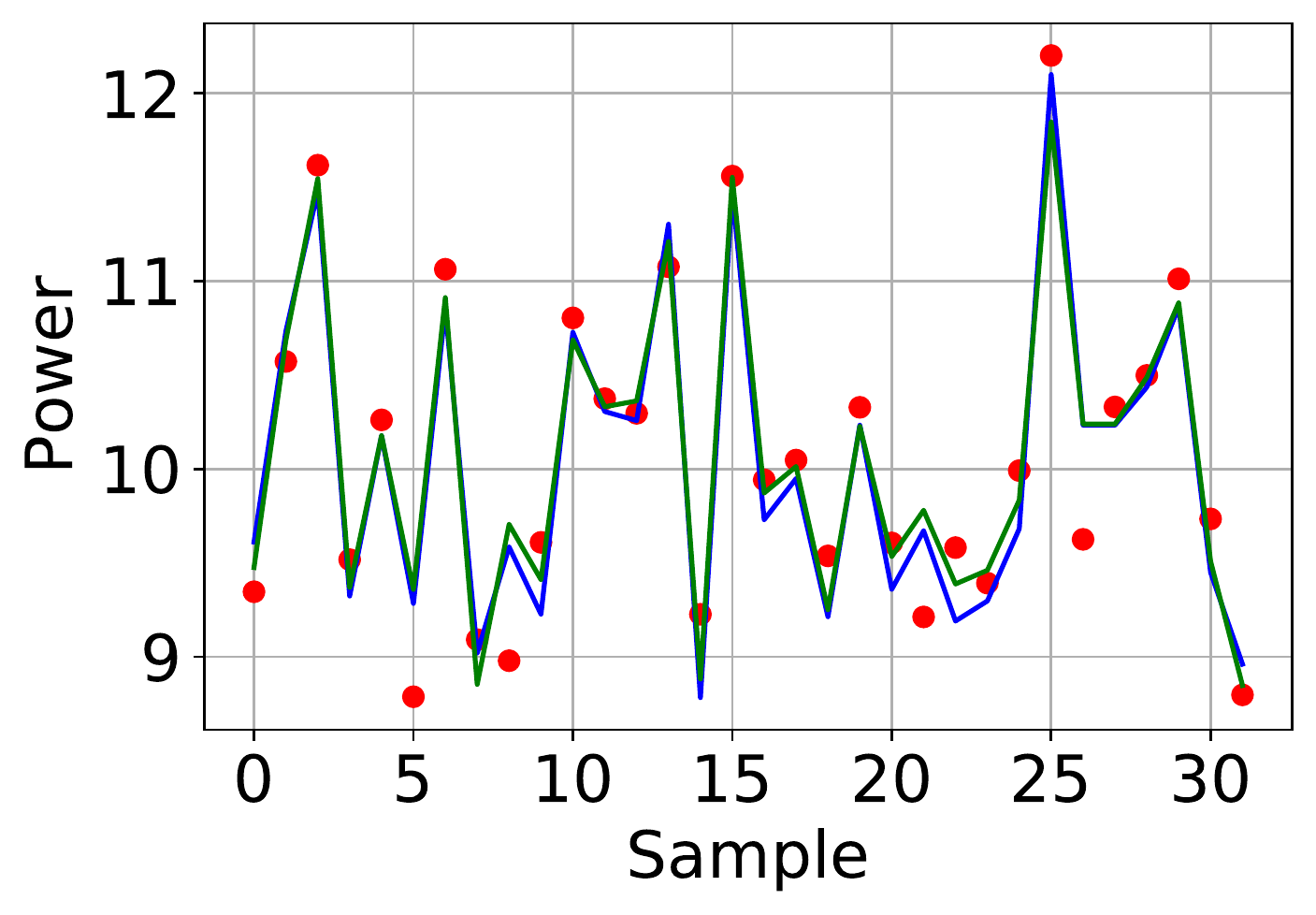}
         \caption{}\label{fig:scheduler_only_prediction_NUC2}
     \end{subfigure}
     \begin{subfigure}[b]{0.24\textwidth}
         \includegraphics[trim={0 0 0 0},clip,scale=0.3]{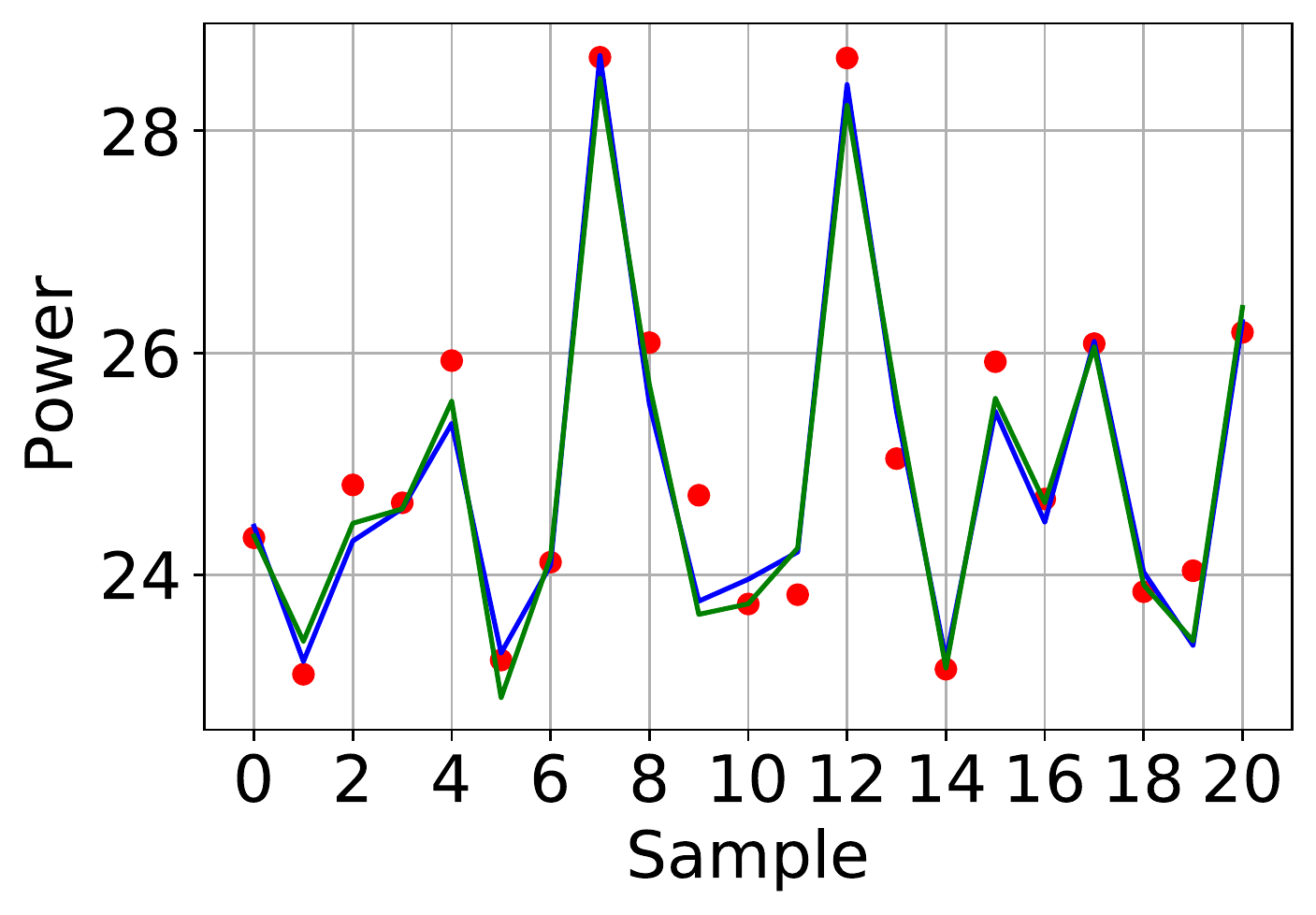}
         \caption{}\label{fig:scheduler_only_prediction_server1}
     \end{subfigure}
     \begin{subfigure}[b]{0.24\textwidth}
         \includegraphics[trim={0 0 0 0},clip,scale=0.3]{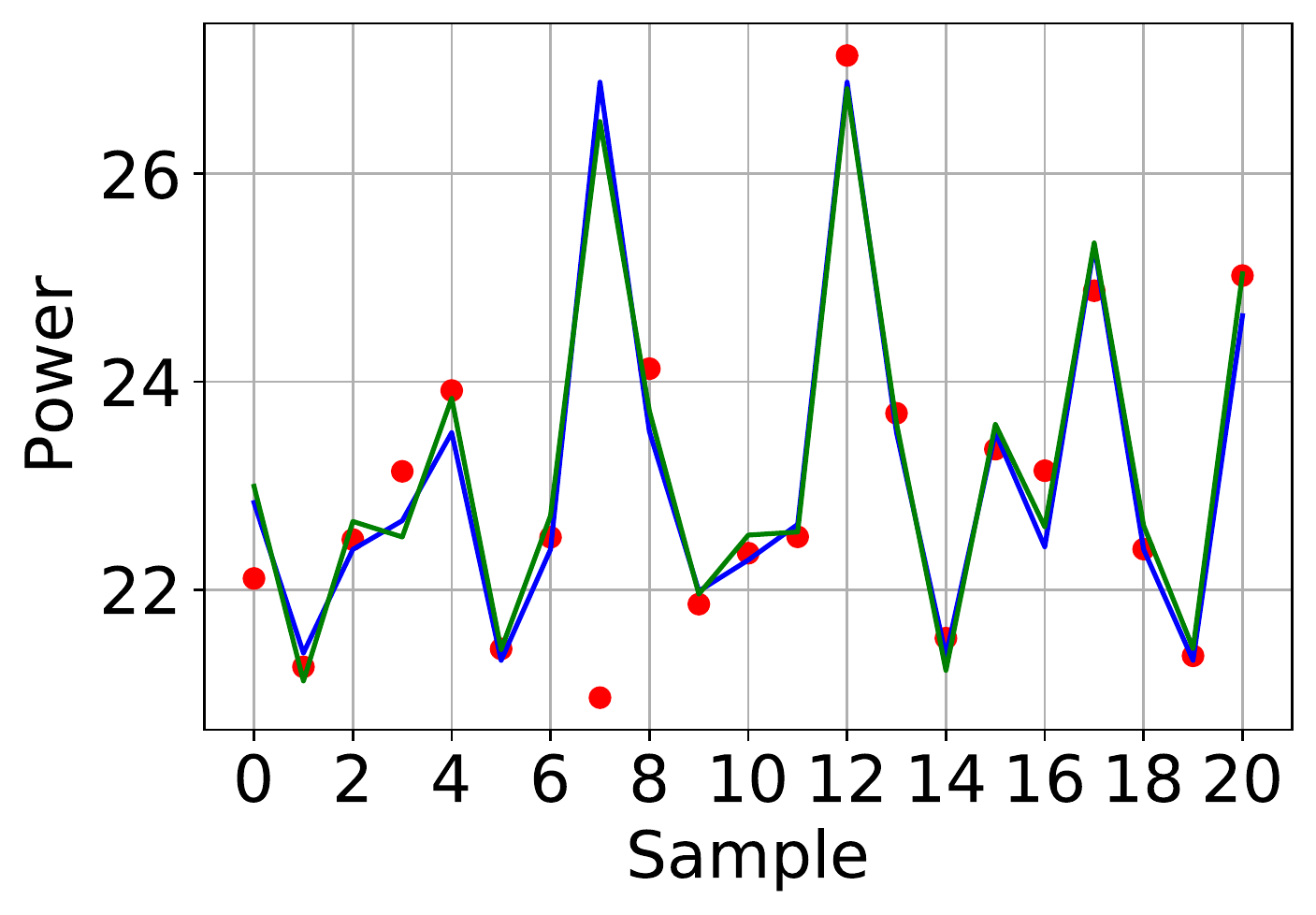}
         \caption{}\label{fig:scheduler_only_prediction_server2}
     \end{subfigure}
        \caption{\textcolor{black}{Testing power consumption samples (red dots) and predictions of the regression model (blue) and the \ac{nn} (green). Training was done with the training data from the default scheduler dataset and prediction on the test set for each of the used computing platforms (a: NUC1; b: NUC2; c: Server1; d: Server2).}}
        \label{fig:scheduler_only_prediction}
        \vspace{-1em}
\end{figure}

Fig.~\ref{fig:scheduler_only_prediction} shows the measurements of the CPU power consumption (red dots) and the prediction made by the regression model (blue line) and the neural network (green line) for all four computing platforms. The figure clearly shows that both models perform similarly well. It is important to notice the scale of the $y$ axis in each of the sub-figures, since it further supports the point of good performance (i.e. the prediction error for both models is small compared to the measured power consumption). Additionally, it is interesting to see that there is almost a perfect overlap in terms of the predictions for both models, showing that the \ac{nn} had enough data to match the performance of the regression model, which was designed with domain knowledge in mind (feature engineering). 

\begin{figure}
\centering
\includegraphics[trim=0 0 87 0,clip,scale=0.35]{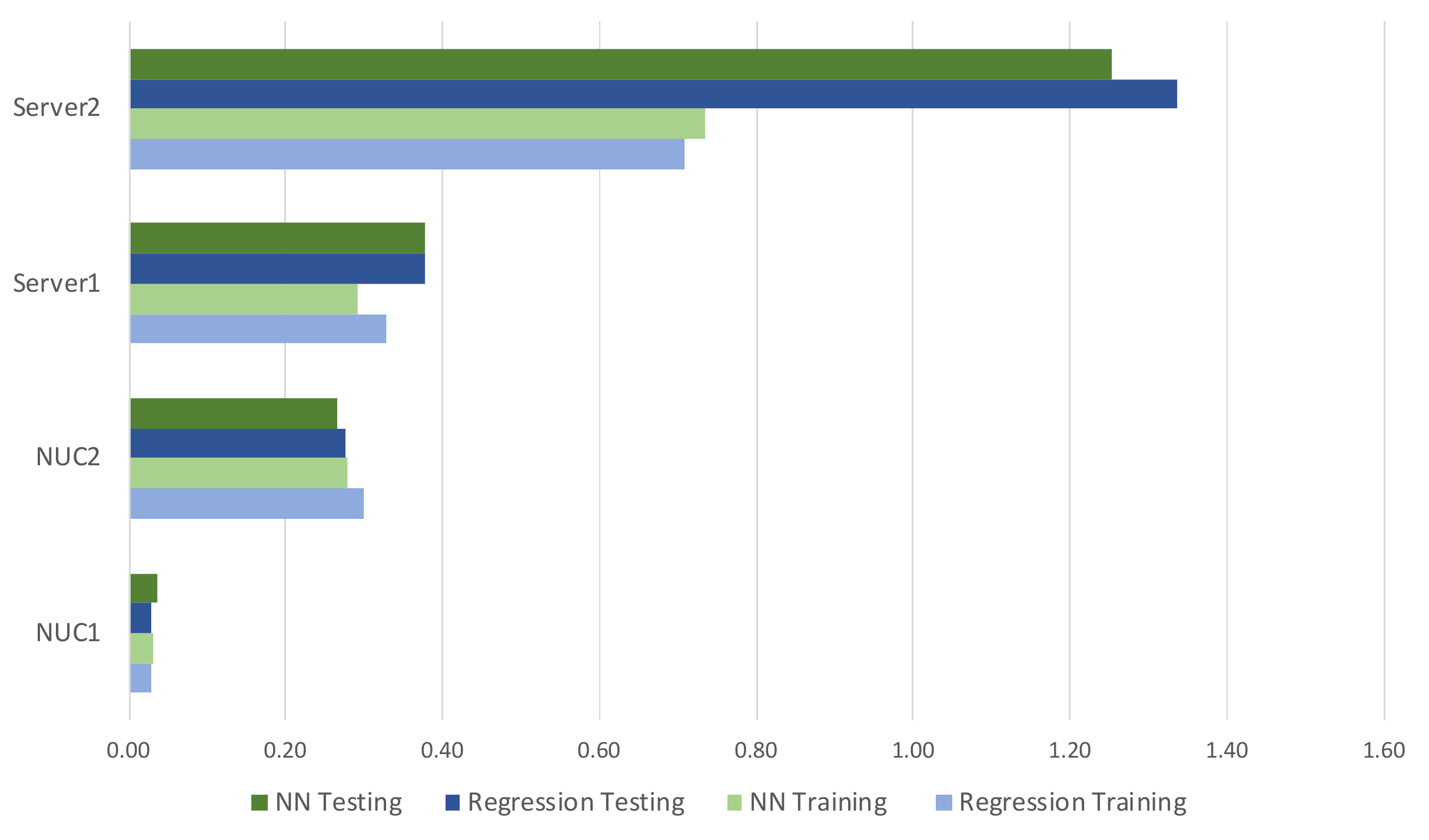}
\caption{\ac{rmse} of the prediction done on the test set from the default scheduler dataset for the regression and \ac{nn} approach trained on the training set of the default scheduler dataset.}
\label{fig:mse_sc_sc}
\vspace{-2em}
\end{figure}

\begin{table}[h]
\caption{Number of training and testing samples per computing platform for the default and custom scheduler datasets.}\label{tab:training_testing_dataset}
\begin{tabular}{l|cc|cc}
\textit{Alias} & Train Default & Test Default & Train Custom & Test Custom \\ \hline
NUC1           & 479        & 119       & 3964           & 991           \\
NUC2           & 128        & 32        & 873            & 218           \\
Server1        & 86         & 21        & 975            & 243           \\
Server2        & 86         & 21        & 564            & 141          
\end{tabular}
\vspace{-1em}
\end{table}
We compute the \ac{rmse} of the power consumption prediction to further inspect the performance of the two models. As Fig.~\ref{fig:mse_sc_sc} shows the \ac{rmse} is similar for both approaches across all tested computing platforms. It is important to highlight that the \ac{rmse} of the testing dataset depicts the performance of the used prediction model (generalization) and the \ac{rmse} of the training dataset shows how well the model fits the training dataset (convergence). Looking at both values together allows us to identify overfitting, i.e. the case when the model performs great on the training dataset but very bad on the testing dataset. 

As shown in Table~\ref{tab:training_testing_dataset}, the overall number of samples per computing platform in the case of the default scheduler dataset is small. However, since the function that describes the relationship between the input features is not very complex (the results suggest that the regression model describes this relationship very well) the \ac{nn} is not large/deep either. This allows us to train the network with even a small number of samples and match the performance of the regression model.  
\subsection{Mixed (default and custom scheduler) Dataset}\label{subsec:mixed_scheduler-only_and_extended_results}
This subsection focuses on studying the potential of the knowledge obtained by models trained for the default scheduler (in our case the \textit{srsRAN} scheduler) to predict the power consumption of a generic scheduler. A generic scheduler, in this case, represents a scheduler that allows any mapping between the channel quality measurements and the chosen \ac{mcs}. By exploring the whole search space in terms of the combination of input features, we can design schedulers that are optimized for power consumption. 

To evaluate the above mentioned generalization, we study the \ac{rmse} of the power consumption prediction made by a model trained on the whole \textit{default scheduler} dataset and tested on the \textit{testing set} of the custom scheduler dataset. Therefore, the training set in this case consists of $598$, $160$, $107$ and $107$ samples while the testing set consists of $991$, $218$, $243$ and $141$ sample, for the \textit{NUC1, NUC2, Server1 and Server2} computing platform respectively (see Table~\ref{tab:training_testing_dataset}).

\begin{figure}
     \centering
     \begin{subfigure}[b]{0.24\textwidth}
         \includegraphics[trim={0 0 0 0},clip,scale=0.3]{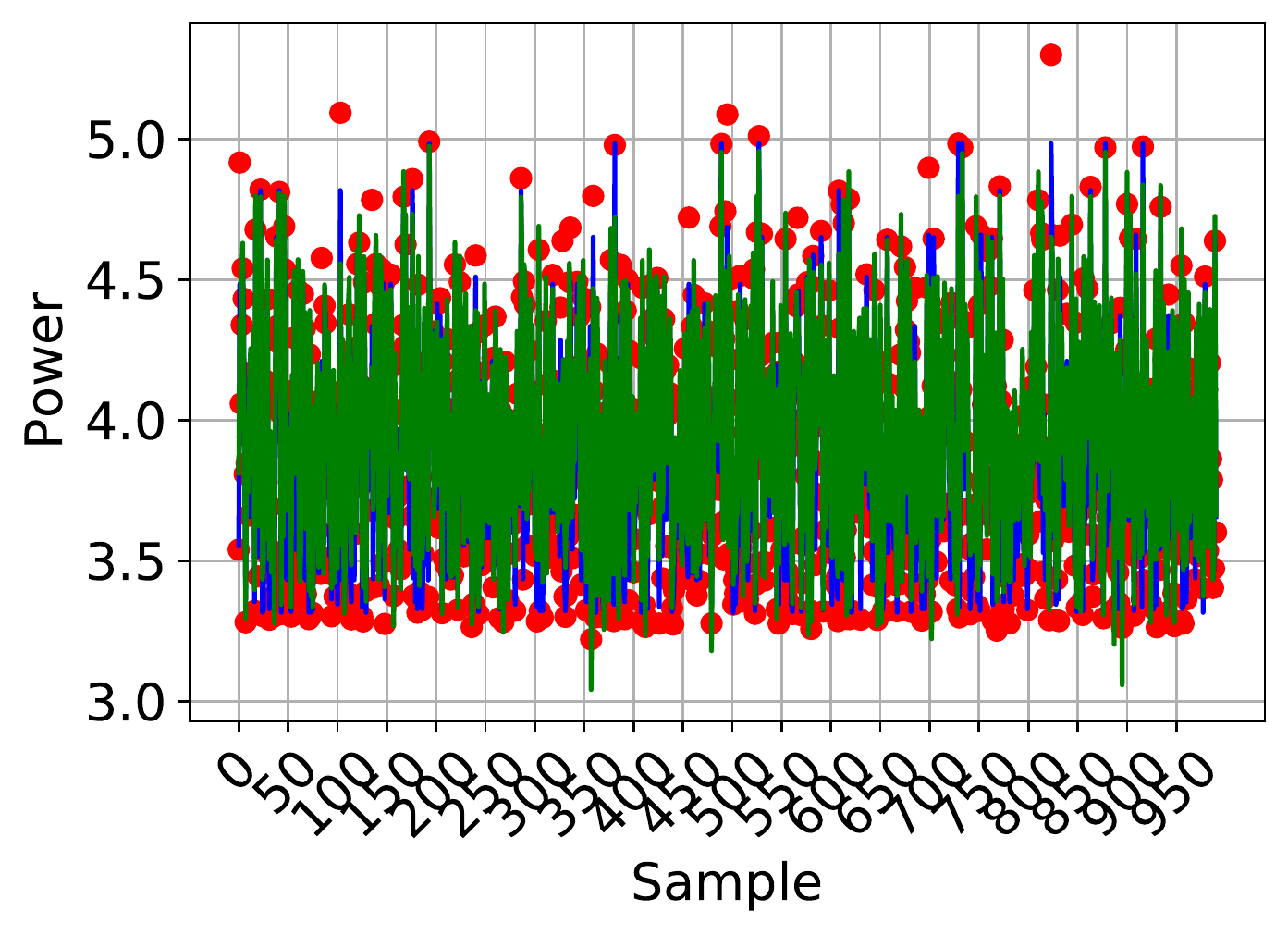}
         \caption{}\label{fig:extended_prediction_NUC1}
     \end{subfigure}
     \begin{subfigure}[b]{0.24\textwidth}
         \includegraphics[trim={0 0 0 0},clip,scale=0.3]{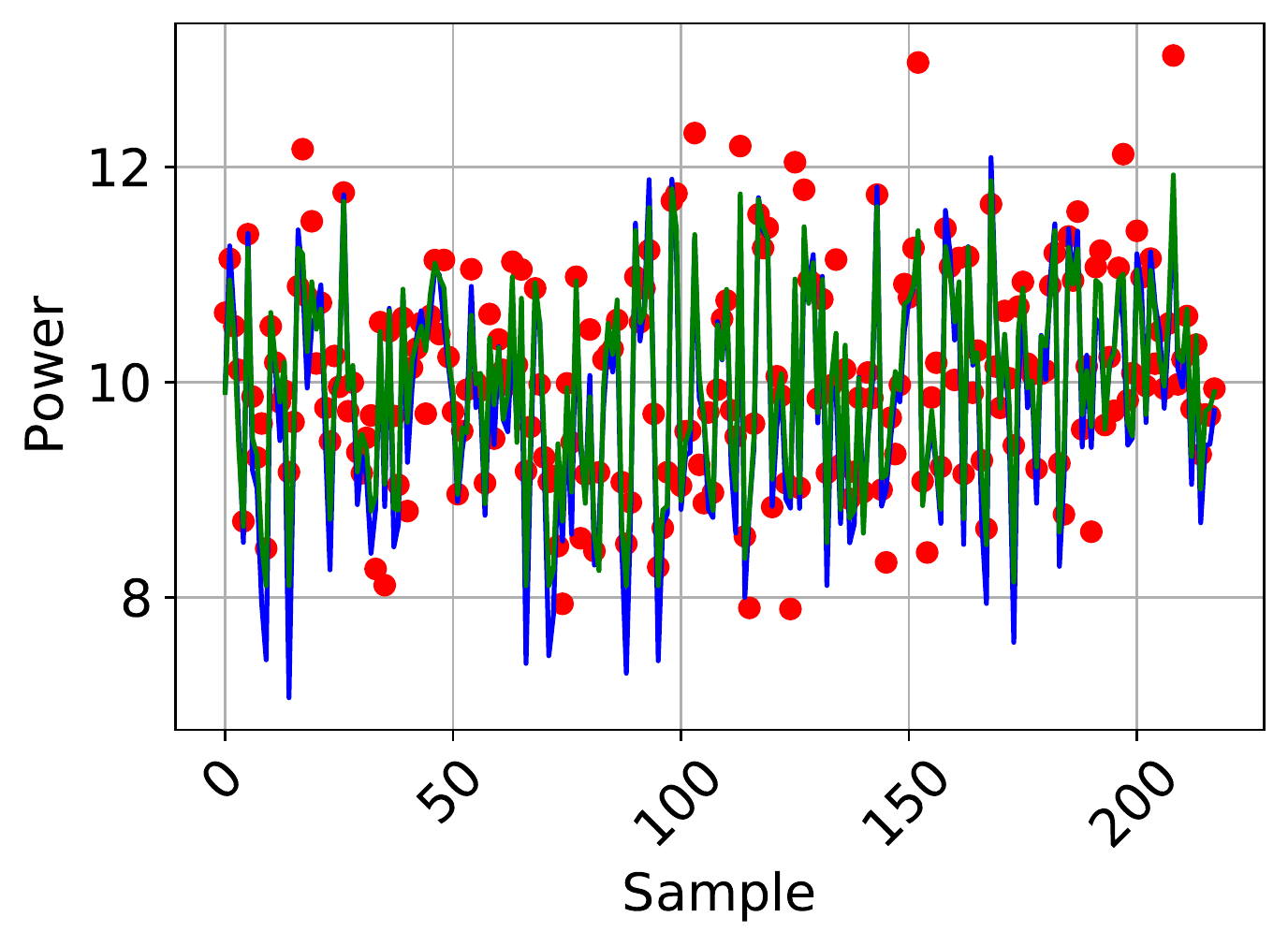}
         \caption{}\label{fig:extended_prediction_NUC2}
     \end{subfigure}
     \begin{subfigure}[b]{0.24\textwidth}
         \includegraphics[trim={0 0 0 0},clip,scale=0.3]{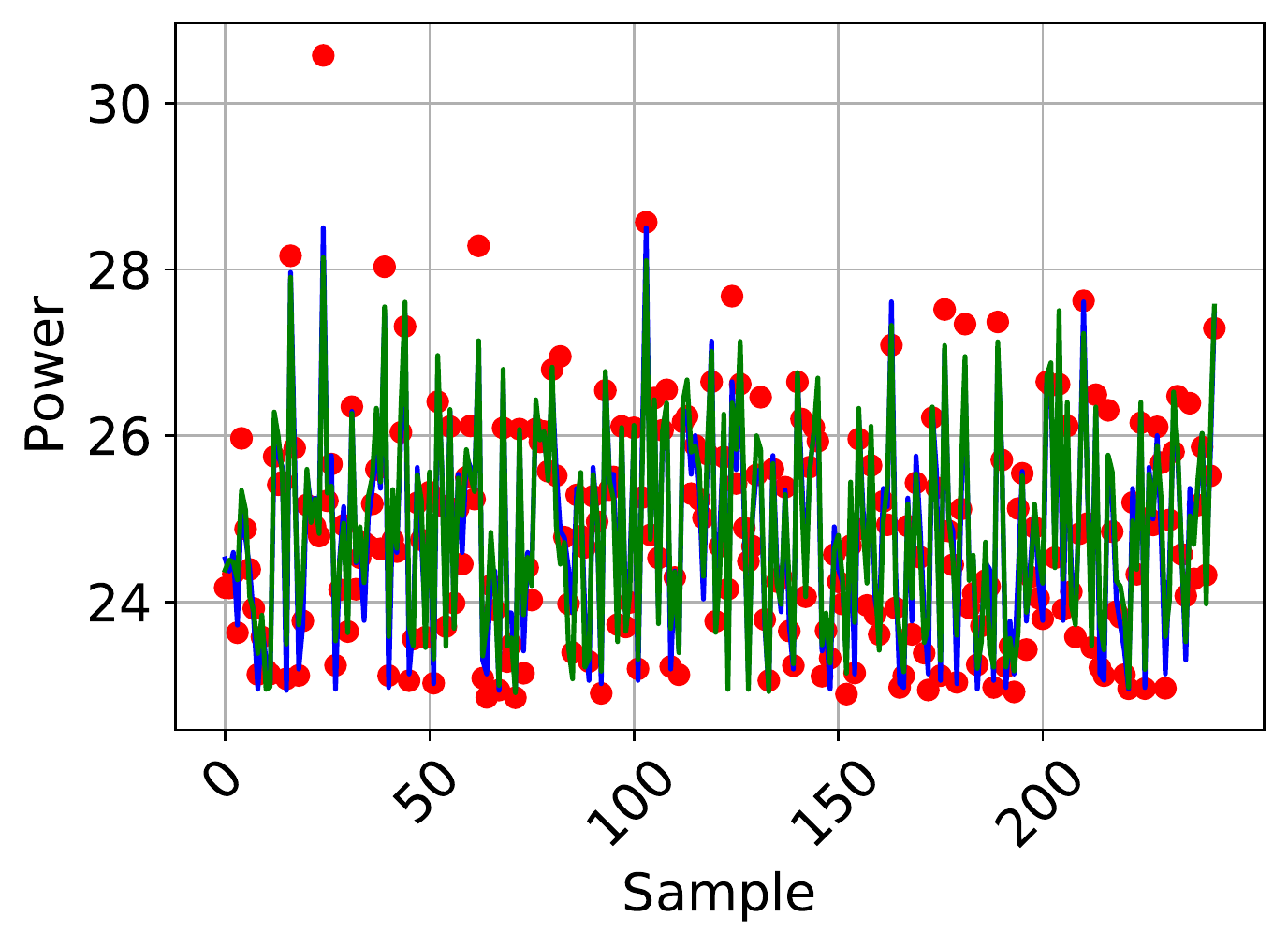}
         \caption{}\label{fig:extended_prediction_server1}
     \end{subfigure}
     \begin{subfigure}[b]{0.24\textwidth}
         \includegraphics[trim={0 0 0 0},clip,scale=0.3]{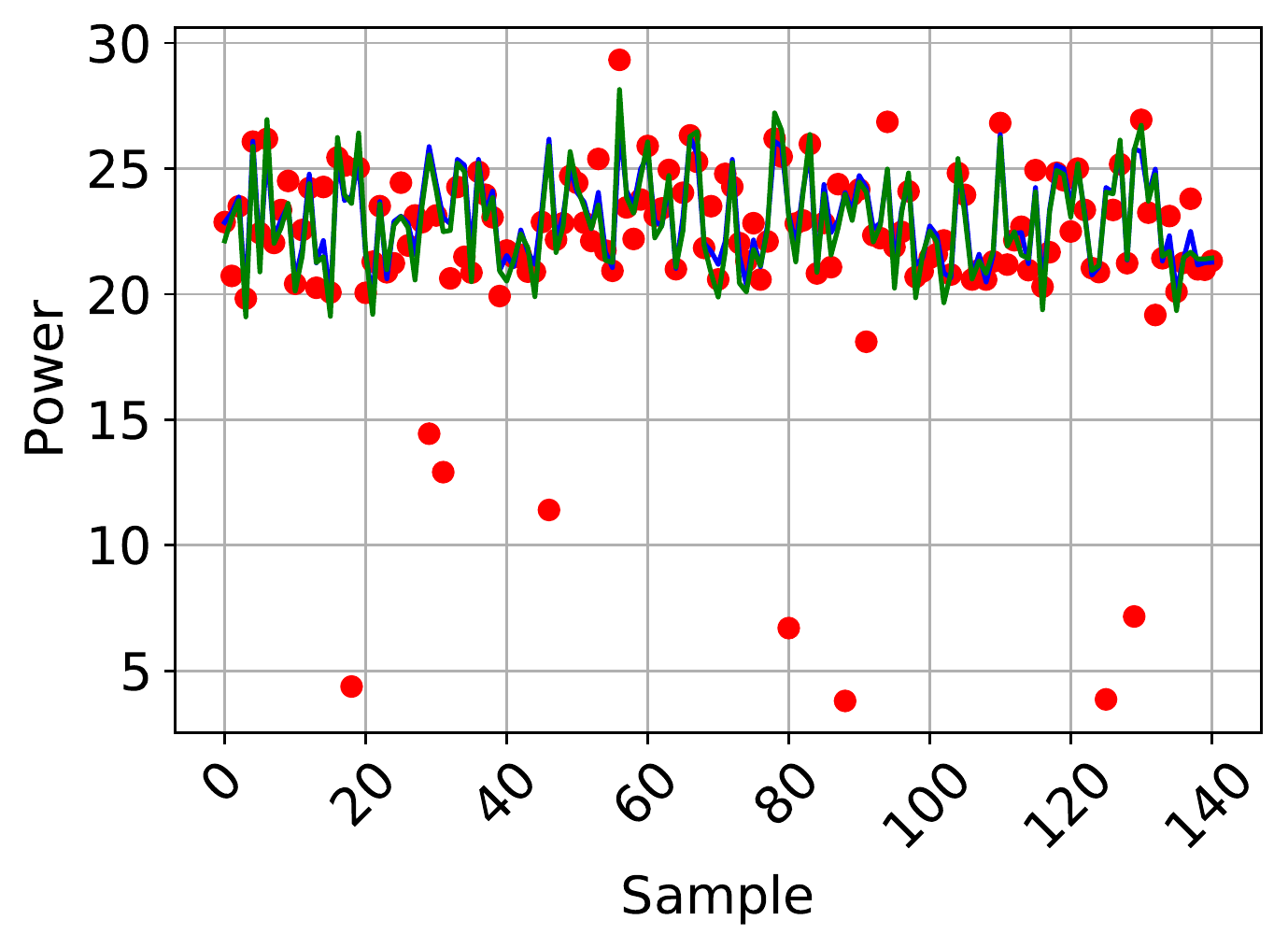}
         \caption{}\label{fig:extended_prediction_server2}
     \end{subfigure}
        \caption{\textcolor{black}{Testing power consumption samples (red dots) and predictions of the regression model (blue) and the \ac{nn} (green). Training done with all default scheduler data and prediction on the test set from the custom scheduler dataset for each of the used computing platforms (a: NUC1; b: NUC2; c: Server1; d: Server2).}}
        \label{fig:sch_train_extended_prediction}
        \vspace{-2em}
\end{figure}

\begin{figure}
\centering
\includegraphics[trim=0 0 0 0,clip,scale=0.35]{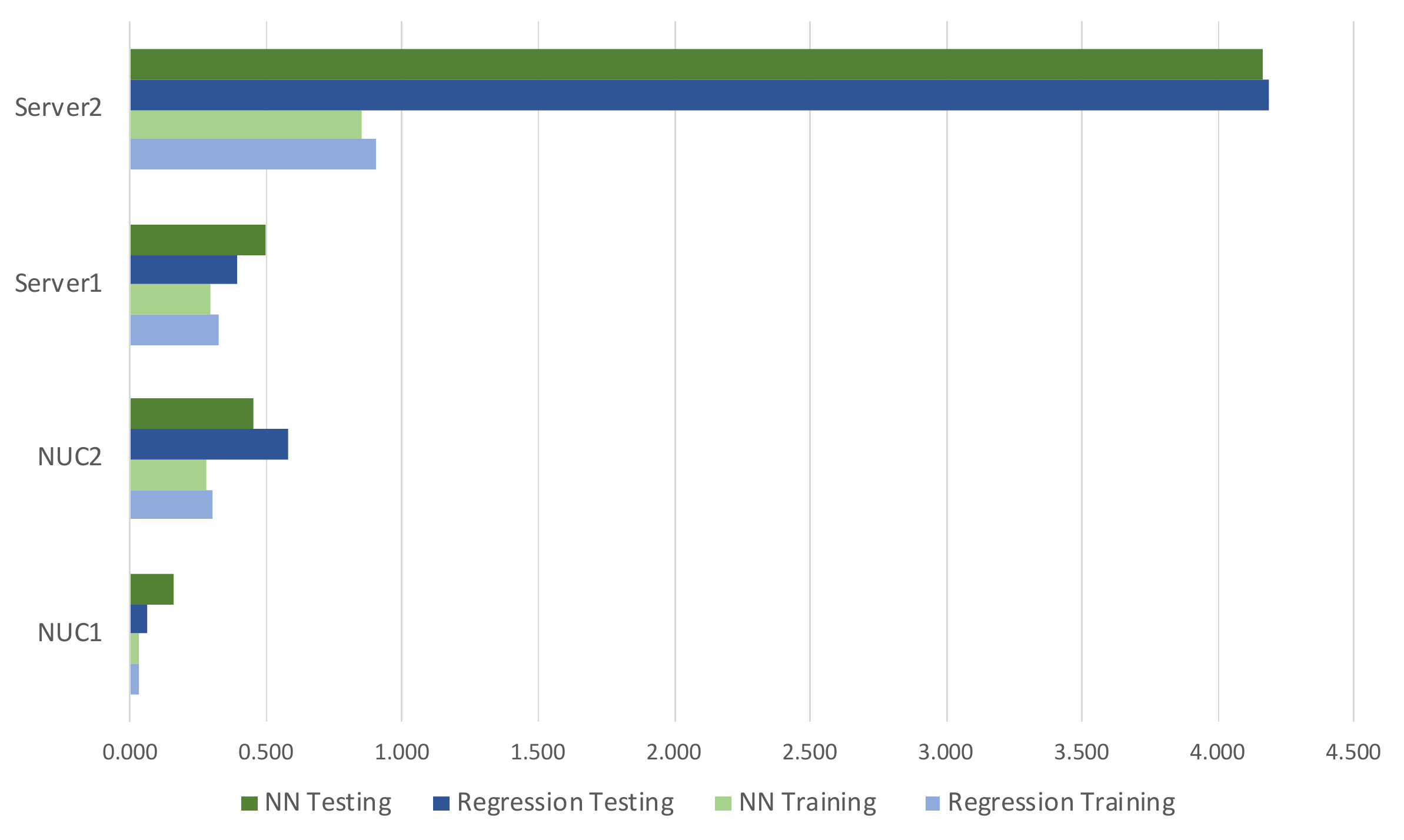}
\caption{\ac{rmse} of the prediction done on the test set from the custom scheduler dataset for the regression and \ac{nn} approach trained on the whole default scheduler dataset.}
\label{fig:mse_sc_extended}
\vspace{-1.3em}
\end{figure}

Fig.~\ref{fig:sch_train_extended_prediction} shows the measurements of the CPU power consumption (red dots) and the prediction made by the regression model (blue line) and the neural network (green line) trained on the whole default scheduler dataset and tested on the testing set of the custom scheduler dataset for all four computing platforms. Similar to Fig.~\ref{fig:scheduler_only_prediction}, the predictions made by both models perform very well on this testing set. It is important to notice that the custom scheduler dataset includes some outlier measurements. For example, Fig.~\ref{fig:extended_prediction_server2} clearly shows that the majority of the measurements are grouped between $20W$ and $26W$. At the same time, the dataset also contains power measurements as small as $5W$. Considering that only a few of the measurements take on such small values, they can be considered outliers of the dataset. As shown in  Fig.~\ref{fig:extended_prediction_server2} none of the two models is able to predict those values. Considering that we rely on \ac{rmse} to measure the performance of the models, it is important to highlight that since the errors are squared before they are averaged, the \ac{rmse} gives a relatively high weight to large errors. This results in a high overall \ac{rmse} in case outliers are detected (see Fig.~\ref{fig:mse_sc_extended}).

Fig.~\ref{fig:mse_sc_extended} shows that the \ac{rmse} is again similar for both models. It is important to notice that in this case we use a different regression model (see equation~\eqref{eq:regression_model_2}), that was specifically built for the custom scheduler dataset. Hence, the regression model has an advantage over the \ac{nn}, since it is specifically built for the testing set in question. However, since the fitting of the model was done with default scheduler data, it was not enough to provide proper generalization. The values in Fig~\ref{fig:mse_sc_extended} are higher compared to the values in Fig.~\ref{fig:mse_sc_sc}, indicating that the default scheduler dataset does not cover enough of the feature space to perform well on the generic scheduler. 

\subsection{Custom Scheduler Dataset}\label{sub:extended_dataset_results}
In this subsection, we will investigate the performance of the models when trained and tested on the custom scheduler dataset. This will allow us to better understand the values studied in the previous subsection (Subsection~\ref{subsec:mixed_scheduler-only_and_extended_results}). 

In this case, the models are trained on the training set of the custom scheduler dataset and tested on the same testing dataset as the models used for the mixed dataset. Table~\ref{tab:training_testing_dataset} shows the number of samples in the training and testing set. 

\begin{figure}
\centering
\includegraphics[trim=0 0 0 0,clip,scale=0.35]{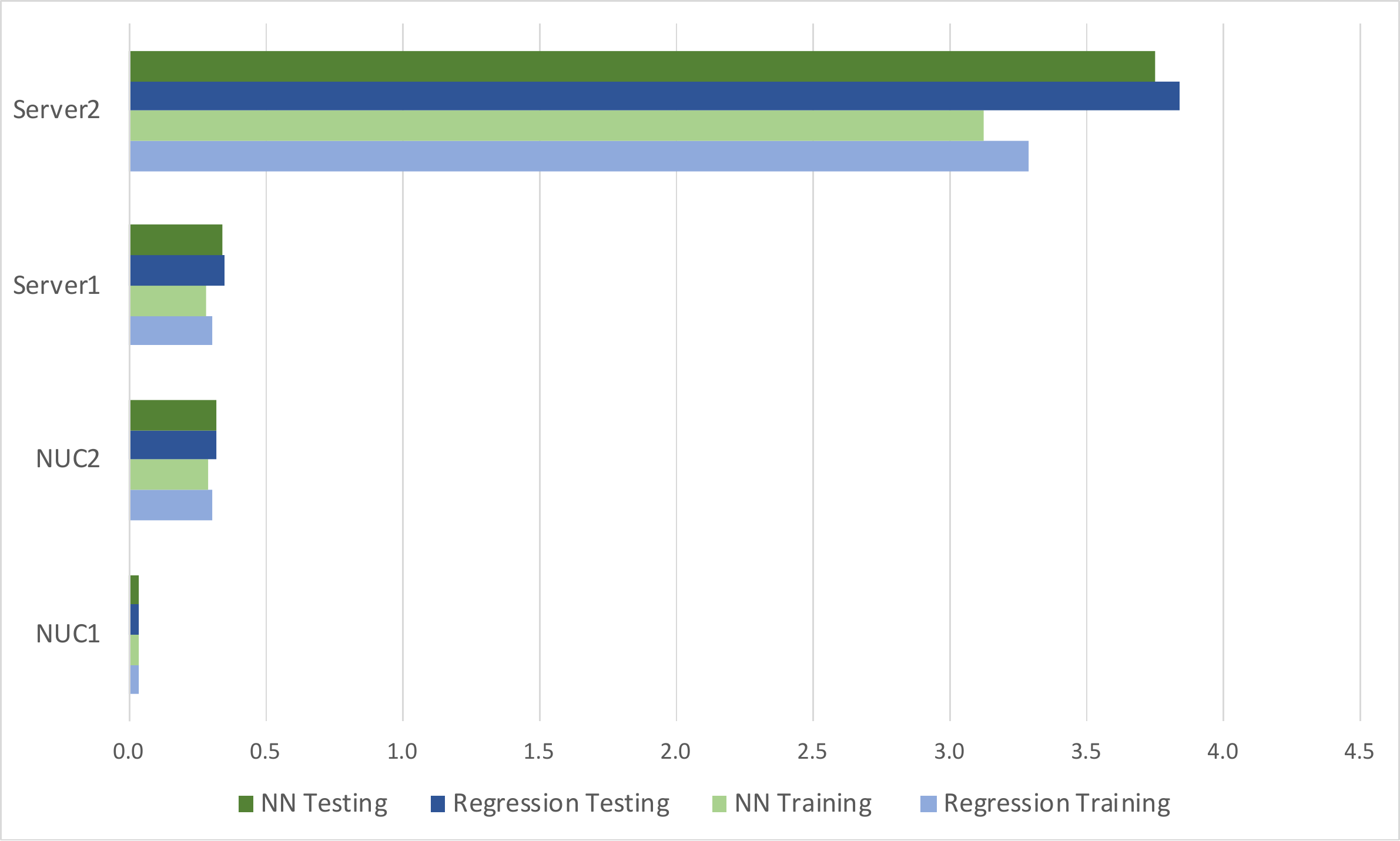}
\caption{\ac{rmse} of the prediction done on the test set from the custom scheduler dataset for the regression and \ac{nn} approach trained on the training set from the custom scheduler dataset.}
\label{fig:mse_extended_only}
\vspace{-2em}
\end{figure}

Considering that the figure comparing the prediction of the two models to the measured values looks almost the same compared to Fig.~\ref{fig:sch_train_extended_prediction}, the new figure is not shown in this paper. However, an important fact to conclude from it is that the models trained on the default scheduler dataset indeed generalize very well. To make sure that the prediction is not degraded we compute the \ac{rmse} for these two models as well (see Fig.~\ref{fig:mse_extended_only}). The figure shows that the \ac{rmse} improves only slightly compared to Fig.~\ref{fig:mse_sc_extended}, again confirming that the models trained on the default scheduler dataset perform very well on the custom scheduler dataset. This leads to an important conclusion showing that a model trained and designed for a specific scheduler can be reused for other types of schedulers as well. Fitting the parameters of the regression model built for the custom scheduler dataset on the new dataset will improve its performance. The same architecture of the \ac{nn} that was used for the default scheduler dataset can be reused for other types of schedulers as well, and further training on the custom scheduler dataset does improve the performance slightly. 
Finally, we tested transfer learning with the black-box model to further improve its performance, but the results suggested that our approach cannot be further improved on these datasets. 

\section{Conclusions}\label{sec:conclusions}
In this work we study the effect that different radio schedulers have on the power consumption of \acp{vbs}. We also proposed a black-box (\ac{nn}) model to predict the said power consumption function and compare it to a previously known regression model. The results show that the black-box model is able to perform as well as the regression model, using the same data, despite no knowledge of the system operation. This suggests that such black-box (\ac{nn}) models provide an advantage over a regression approach in situations where domain knowledge is not available or is hard to acquire. 
We have also provided a detailed mapping of the data collection, model training, and inference, as well as the deployment of control policies on the \ac{oran} architecture. 


\section{Acknowledgments}
We would like to say thank you to Diarmuid Collins for the help in the use and discussions related to \textit{srsRAN}.
This material is based upon works supported by the Science Foundation Ireland under Grants No. 17/CDA/4760 and 13/RC/2077\_P2.
\vspace{-0.5em}


%

%



\ifCLASSOPTIONcaptionsoff
  \newpage
\fi

\bibliographystyle{IEEEtran}
\bibliography{main.bib}

\end{document}